\documentclass[letterpaper, 10pt, journal]{IEEEtran}

\usepackage[utf8]{inputenc}
\usepackage[T1]{fontenc}
\usepackage{adjustbox}
\usepackage{amsmath}
\usepackage{amsfonts}
\usepackage{amssymb}
\usepackage{graphicx}
\usepackage{siunitx}
\usepackage{color}
\usepackage{xcolor}
\usepackage{booktabs}
\usepackage{dblfloatfix}
\usepackage{url}
\usepackage{nth}
\usepackage{hyperref}
\hypersetup{
    colorlinks=true,
    linkcolor=black,  
    citecolor=black,
    urlcolor=black,
}

\usepackage{subfig}
\usepackage{hyphenat}
\usepackage{cite}

\title{Human-Piloted Drone Racing: \\Visual Processing and Control}

\author{Christian Pfeiffer and Davide Scaramuzza

\thanks{Manuscript received: October 15, 2020; Revised: January 21, 2021; Accepted: February 24, 2021.}

\thanks{This paper was recommended for publication by Editor Pauline Pounds upon evaluation of the Associate Editor and Reviewer's comments.}

\thanks{The authors are with the Robotics and Perception Group, Department of Informatics, University of Zurich, and Department of Neuroinformatics, University of Zurich
and ETH Zurich, Switzerland (\protect\url{http://rpg.ifi.uzh.ch}).}

\thanks{This work was supported by the Ernst Göhner Foundation and University of Zurich Alumni Fonds zur Förderung des Akademischen Nachwuchses (FAN Fellowship), by  the National Centre of Competence in Research (NCCR) Robotics through the Swiss National Science Foundation (SNSF) and the European Union’s Horizon 2020 Research and Innovation Programme under grant agreement No. 871479 (AERIAL-CORE) and the European Research Council (ERC) under grant agreement No. 864042 (AGILEFLIGHT).}

\thanks{Digital Object Identifier (DOI): see top of this page.}
}

\begin{document}

\maketitle

\begin{abstract}
Humans race drones faster than algorithms, despite being limited to a fixed camera angle, body rate control, and response latencies in the order of hundreds of milliseconds. A better understanding of the ability of human pilots of selecting appropriate motor commands from highly dynamic visual information may provide key insights for solving current challenges in vision-based autonomous navigation. This paper investigates the relationship between human eye movements, control behavior, and flight performance in a drone racing task. We collected a multimodal dataset from $21$ experienced drone pilots using a highly realistic drone racing simulator, also used to recruit professional pilots. Our results show task-specific improvements in drone racing performance over time. In particular, we found that eye gaze tracks future waypoints (i.e., gates), with first fixations occurring on average $1.5$ seconds and $16$ meters before reaching the gate. Moreover, human pilots consistently looked at the inside of the future flight path for lateral (i.e., left and right turns) and vertical maneuvers (i.e., ascending and descending). Finally, we found a strong correlation between pilots’ eye movements and the commanded direction of quadrotor flight, with an average visual-motor response latency of $220$ ms. These results highlight the importance of coordinated eye movements in human-piloted drone racing. We make our dataset publicly available.
\end{abstract}


\section*{Dataset}
The dataset can be downloaded at \protect\url{https://osf.io/gvdse/}.

\section{Introduction}
\IEEEPARstart{F}{irst}-person view (FPV) drone racing has become a popular televised sport in recent years.
FPV pilots observe a visual display showing a live stream video from a drone-mounted camera and use a hand-held remote for sending collective thrust and body rate commands in order to control six degrees-of-freedom (DoF) motion of the drone in three-dimensional space. Various drone racing formats exist in which pilots fly alone (i.e., time trials), alternatingly (e.g., endurance race), or simultaneously with other pilots. Racing pilots compete for completing a predefined, often three-dimensional obstacle course as fast as possible. To achieve the visual-motor coordination skills required for these top-level performances, human pilots often require years of training, which has led some passionate individuals to pursue drone racing as a full-time profession. A better understanding of the underlying processes enabling human pilots to successfully complete drone races, and in particular the process of selecting appropriate motor commands from highly dynamic visual information may provide key insights for solving current challenges in state estimation and planning for vision-based autonomous navigation.

\begin{figure}[t]
    \centering
    \includegraphics[width=\linewidth,trim={0 20 0 0},clip]{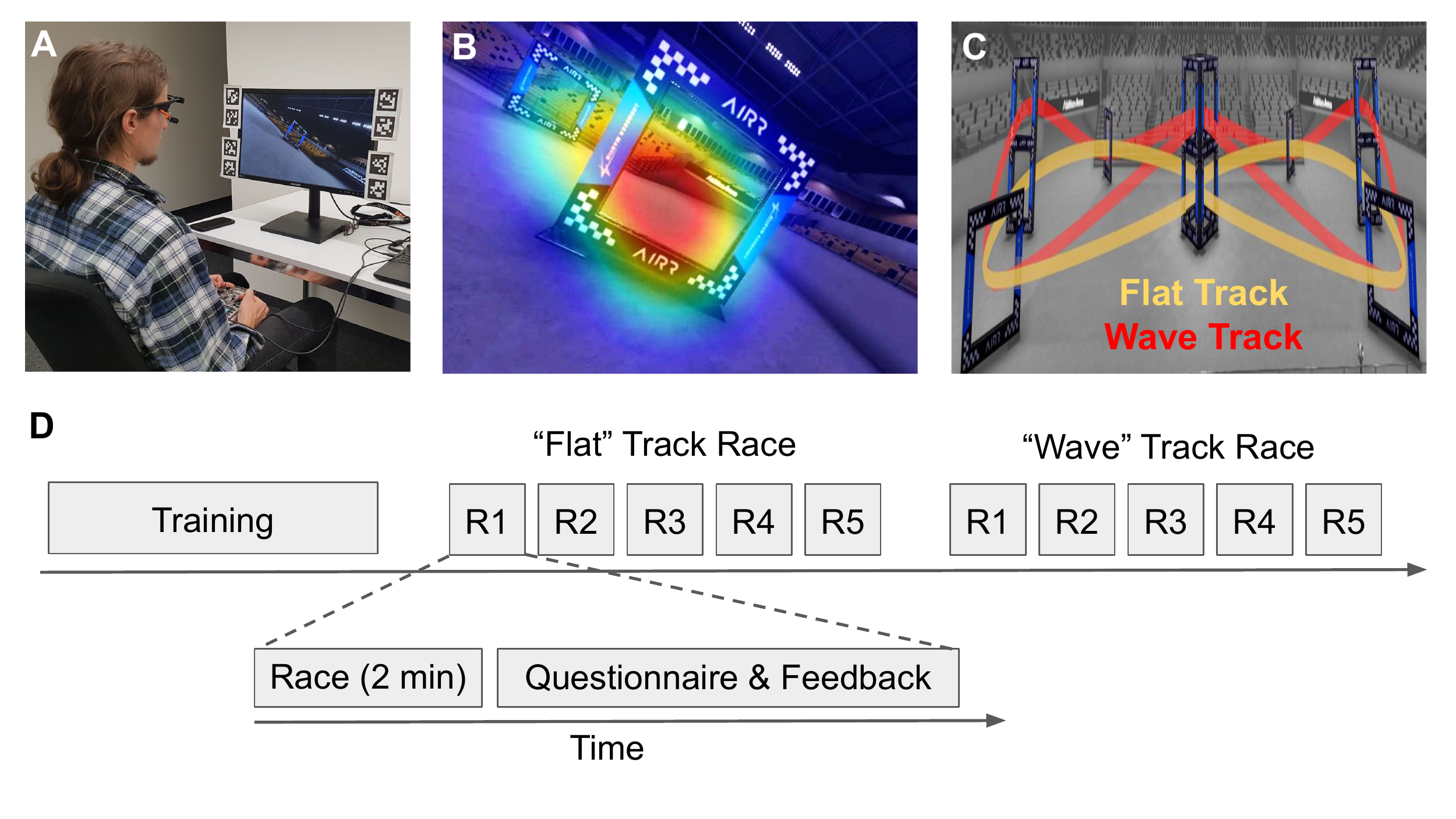}
    \caption{(A) Experimental setup showing a participant equipped with eye tracking glasses, a remote controller, and a first-person view (FPV) display. (B) Exemplar gaze fixation heatmap for a left turn of the "Flat" race track. (C) Overview of race tracks, showing the gates (in blue) and exemplar flight paths for the "Flat" track (in yellow) and the "Wave" track (in red). (D) Timeline of the experimental procedure, including training and two races of five runs (R1-R5). The run procedure is shown in the lower part of the panel.}
    \label{fig:fig1}
\end{figure}


An important step towards this goal is to extend knowledge about the relationship between eye movements, quadrotor control, and flight performance in humans during drone racing tasks. We thus collected a multimodal dataset consisting of eye tracking, control command, and drone state data from $\si{21}$ experienced drone pilots using a highly realistic drone racing simulator, and performed an analysis of flight performance, quadrotor control, and eye movement data.

\section{Related Work}
 
Human-drone interaction has been studied in various contexts, including control interface design \cite{Rognon2019HapticDrone, Cherpillod2019EmbodiedDrone}, shared autonomy applications \cite{Sheridan2016Human-RobotInteraction}, and brain-computer interfaces \cite{Rosca2018QuadcopterBCI}. However, FPV drone racing, and in particular the visual-motor coordination processes involved in fast and agile flight have received little attention by the research community.

The methods for experimentally testing visual-motor coordination in this study are adapted from previous work on car driving. \cite{Land1994WhereSteer} were the first to report that when negotiating road intersections or following curved roads, human drivers focus their eye gaze onto specific points of the road before initiating a steering wheel motion in the same direction. These gaze fixation on so-called tangent-points (i.e., eye gaze directed towards the inner bend of the curve) and future-points (i.e., eye gaze directed to parts of the road of the future driving path) have been repeatedly observed in both in real-world \cite{Boer1996TangentNegotiation, Lappi2013BeyondDriving} and simulator-based car driving studies \cite{Authie2011OptokineticDriving, Negi2019DifferencesStudy}. More recent work \cite{Land1995WhichSteering, Okafuji2018SteeringControl} suggests that not all visual information is necessary for car driving but that two points---one far from the car close to the horizon, and one close to the car---are sufficient for producing accurate lane following behavior. Removal of either the far or the close view impairs driving performance in a simulator\cite{Land1995WhichSteering, Salvucci2004ASteering, VanLeeuwen2014VerticalEvaluation}. Another relevant observation \cite{Tuhkanen2019HumansSteering} is that tangent and future point fixations occur at fixed time lags relative to steering commands, between $300$ ms and $5$ sec in advance, depending on the tested driving maneuver and test conditions \cite{Marple-Horvat2005PreventionPerformance, Wilson2008EyePerformance}. The relationship between gaze behavior and control commands in car driving can be empirically tested by computing the horizontal angle between eye gaze direction and the car's forward direction (i.e., gaze angle) and the rotation angle of the steering wheel (i.e., steering angle). \cite{Marple-Horvat2005PreventionPerformance, Wilson2008EyePerformance} used cross-correlation analysis on these metrics and found strong correlations (r=$0.7$) at a temporal lag of $300-400$ ms, indicating that gaze behavior precedes the steering motion \cite{Marple-Horvat2005PreventionPerformance}. Extending these works to car racing, \cite{vanLeeuwen2017DifferencesEye-tracking} compared the performance of professional racing car drivers to non-professional drivers. These authors found that racing car drivers used overall higher throttle inputs, drove corner-cutting trajectories, and controlled brake pedal inputs more dynamically than non-professional drivers. Racing car drivers' gaze fixations also showed more exploratory eye movements beyond tangent point fixations \cite{vanLeeuwen2017DifferencesEye-tracking}. These studies provide compelling evidence for the ability of humans to navigate cars across winding roads at various speeds using specific spatio-temporal patterns of eye movements and control inputs. 

It is worth noting that car driving and FPV drone racing differ in terms of perception, control, and planning aspects. Most importantly, car drivers use only two control inputs (i.e., steering wheel rotation, accelerator/brake pedal) for controlling two DoFs (i.e. yaw rotation and linear forward-backward motion) of the car relative to the road surface. By contrast, drone pilots simultaneously use four input commands (i.e., collective thrust and body rates), for controlling six DoFs of the drone. Drone pilots also have to actively control the elevation level of the drone relative to the ground floor and ceiling, which is not required in typical car driving scenarios. Drone pilots use monocular cameras with limited field of view, whereas car drivers have access to wide field-of-view stereo vision. These perception and control differences raise the question to which degree visual-motor coordination principles differ between car driving and drone racing.

Although research on human drone racing is sparse, autonomous drone racing has recently become a topic of intensive research in robotics. Questions addressing state estimation, autonomous navigation, and time-optimal planning have motivated competitions, such as the real-world IROS Drone Racing Competitions (from $2016$ to $2019$)~\cite{Moon16ram,Moon19springer} and the $2019$ AlphaPilot Drone Racing Challenge~\cite{alphapilot}, but also simulator-based competitions, such as the NeurIPS $2019$ Game of Drones competition~\cite{gameofdrones}, and the UZH-FPV Drone Racing VIO competitions (from $2019$ to $2020$)~\cite{Delmerico2019icra}. The aim of this research spanning across computer vision, robotics, and machine learning is developing better algorithms that can cope with unreliable state estimation and low-latency perception and action cycles and time-optimal trajectory planning combining machine learning methods with classical robotics methods for perception and navigation, as well as systems integration. Drone racing tasks thus provide the unique opportunity for directly comparing the performance of autonomous drones and human-piloted drones on the same race tracks using identical test conditions.

\section{Contributions}
The main contributions of this work are: ($1$) We found that human drone racing pilots consistently direct their eye gaze on future waypoints in the direction of the future flight trajectory, both for lateral maneuvers (left and right turn) and for vertical maneuvers (ascending and descending trajectory). ($2$) We found that gaze fixations on upcoming gates occurred up to $1.5$ sec and $16$ meters before the drone reached the gates. ($3$) We found a strong correlation between pilots’ eye movements and the commanded direction of quadrotor flight, with an average visual-motor response latency of $220$ ms. ($4$) We publicly release our dataset, consisting of eye movement, control command, and drone state ground truth data from typical drone racing maneuvers.

\section{Methods}

\subsection{Participants}
Twenty-one right-handed male volunteers with a mean age of $30$ years and an age range of $18$-$42$ years participated in this study. All participants were recruited online from a website of a local drone racing association. All participants had prior experience with drone racing, took part in two or more official drone races, and had at least one year of experience in FPV flight. The study protocol was approved by the local Ethical Committee of the University of Zurich and the study was conducted in line with the Declaration of Helsinki. All participants gave their written informed consent before participating in the study and received monetary compensation of $25$ Swiss Francs per hour.

\subsection{Apparatus and Stimuli}
The experiment took place in a normal lit quiet room at the University of Zurich. The participant was comfortably seated in front of a computer monitor (Dell U2419HC, $53$$\times$$27$ cm screen size, $1920$$\times$$1080$ pixels resolution, 60 Hz refresh rate) at $60$ cm distance resulting in approximately $47$$\times$$27$ degrees field of view. The monitor was used to present the participant with a FPV video the drone simulator. The participant was equipped with a remote controller (FrSky Taranis X9D Plus) connected via USB with the simulator laptop. Four control commands were mapped to the remote controller in Mode $2$ (i.e. left stick: throttle and yaw, right stick: pitch and roll). The participant controlled the gimbal sticks of the remote controller using the thumbs or thumbs and index fingers of both hands. Two laptops were used (i.e., one for running the simulator and logging drone state data, one for recording eye tracking data).

\subsubsection{Eye Tracker}
Participants wore a PupilLabs Glasses eye tracking device (PupilLabs, Berlin), consisting of a world and an eye camera. Eye videos were recorded at $200$ Hz and world video at $50$ Hz. Raw gaze data and eye calibration data were captured with the PupilCapture software (v$1.19$) and saved to the hard drive of the eye tracking laptop. We tracked gaze data of the dominant eye (i.e., assessed with the look-through-card method by Dolman \cite{Gould1910Amethod}), because dominant eye movements are considered to be more accurate than non-dominant eye movements \cite{Ehinger2019A1000}. We thus performed monocular eye tracking, because vergence information from stereo tracking was not required due to the fixed viewing distance of the participant form the computer monitor. Eye tracker calibration was performed at the beginning of the experiment and re-calibration was performed every $15$ min between drone racing experimental recordings. The calibration procedure consisted of two repetitions of the built-in calibration program of the PupilCapture software. A fixation target (i.e. black circle-dot pattern) was shown on a gray background alternatingly at $12$ target locations on the screen for $2$ sec with $1$ sec inter-stimulus interval. The second calibration run was used for validating the calibration quality of the first run. We observed across subjects and calibration runs a median validation accuracy of $0.5$ deg of visual angle for all subjects. During the experiment participants were instructed to keep their head at a fixed position, avoid touching their face and keep their facial muscle relaxed, in order to avoid small displacements of the eye tracker that could induce degradation of the calibration quality.

\subsubsection{Drone Racing Simulator}
We used the AlphaPilot simulator, a modified version of the Drone Racing League simulator, which is used by professional pilots for training, and pilot recruitment, and even online competitions, such as the DRL TryOuts with money prizes of several hundred thousand USD. The quadrotor model was a replica of the racing drone used for the AlphaPilot competition \cite{Foehn2020}, with $3.2$ kg weight, $1.5$ kg maximum thrust per motor, $0.127$ Nm motor torque, and a single front-facing RGB camera with $30$ deg up tilt and $120$ deg field of view. Pilot training and races took place in an indoor environment "Addition Arena'', which has a size of $35$$\times$$20$$\times$$15$ meters. Racing gates had a rectangular shape and an outer diameter of $3$$\times$$3$$\times$$0.25$ meters and an inner opening of $2$$\times$$2$$\times$$0.25$ meters (Fig. \ref{fig:fig1}b). A main focus of this study was testing typical maneuvers observed in drone racing, in particular left-right turn and upward-downward flight maneuvers. We thus designed two race tracks each consisting of $10$ gates positioned in a figure-eight configuration when seen from the top (Fig. \ref{fig:fig3}). This gate configuration has the advantage of being symmetric and allowing to record data from left and right turns continuously in a short time, therefore maximizing the amount of repetitions performed during the experiment. The first track, referred to as "Flat" track, had all gates at the same height (i.e. $1.75$ meter gate-center distance to the ground floor) and can be considered the easier track of the two, because no elevation changes were required for passing the gates. This track was chosen because it mainly required pilots to perform lateral translations at the same altitude, which can be considered more similar to car driving studies. For the second track, referred to as "Wave" track, gates were alternatingly placed at either $1.75$ meter height (for gates $1$, $3$, $6$, $8$) or at $3.5$ meter height (for gates $0$, $2$, $4$, $5$, $7$, $9$), which required the pilots to perform elevation changes when flying the race track. Thus, pilots had to not only navigate along the lateral direction, but had to transition between different altitudes for passing through the gate. This task clearly extends the control demands beyond car driving. This is why racing on the Wave track is more challenging than racing on the Flat track. Ground truth drone state data (i.e., position, rotation, velocity, angular rates) and control commands (i.e., throttle, roll, pitch, yaw) were collected from the UDP interface of the simulator at $500$ Hz and saved as .csv files to the hard drive of the simulation computer. High-resolution Camera images were recorded with the screen capture software Kazam at $800$$\times$$600$ pixels resolution and $60$ Hz.

\subsection{Experimental Procedure}
Upon arrival at the lab, participants answered a questionnaire about their drone experience. Then the experimental procedure and task was explained. This was followed by equipping them with the eye tracking glasses and remote controller, and eye tracker calibration. Participants then performed a training aimed at familiarizing them with the handling of the particular racing drone, camera angle, and body rate settings of the simulator. For this reason, participants performed training flights on different tracks not used for the experiments, where they performed individual flight maneuvers, such as hairpin going around two gates, slalom around $20$ gates, and Split-S maneuver (i.e. passing two gates stacked on top of each other by either passing from the top gate to the bottom gate or vice versa. The total duration of the training block was $25$ min. This was followed by two racing tasks, starting with five runs of the Flat track race, followed by five runs of the Wave track race (Fig. \ref{fig:fig1}d). We aimed to keep the flying experience as close as possible to real-world drone racing, by asking participants to execute repeated runs on the same track to allow them to improve lap times over time. Participants were instructed to complete as many laps within $120$ sec after start as possible. They were told that they should fly as fast as possible without crashing. If pilots crashed into obstacles, the drone was not damaged and the pilots were requested to resume the flight. However, crashes usually slow down the flight progress along the track and were thus discouraged. Participants were told that after each race run, they would be asked to complete a questionnaire evaluating their flight performance and workload required during the race run, and finally receive feedback about the number of laps completed and their fastest lap times of the preceding run. They were also instructed to try and improve their previous performance with every run and allowed breaks between racing runs. Questionnaires consisted of the German version of the NASA-TLX scale \cite{Hart2006NASA-TLX:Later} measuring subjective ratings of workload, and adapted versions of performance assessment and agency ratings \cite{Longo2008WhatApproach} using a $7$-point horizontal visual analogue scale implemented in Google forms. Data was coded so that participant identity was not stored along with questionnaire ratings. The total duration of the experiment was $2$ hours.

\begin{figure}[t]
    \centering
    \includegraphics[width=\linewidth,trim={120 150 180 80},clip]{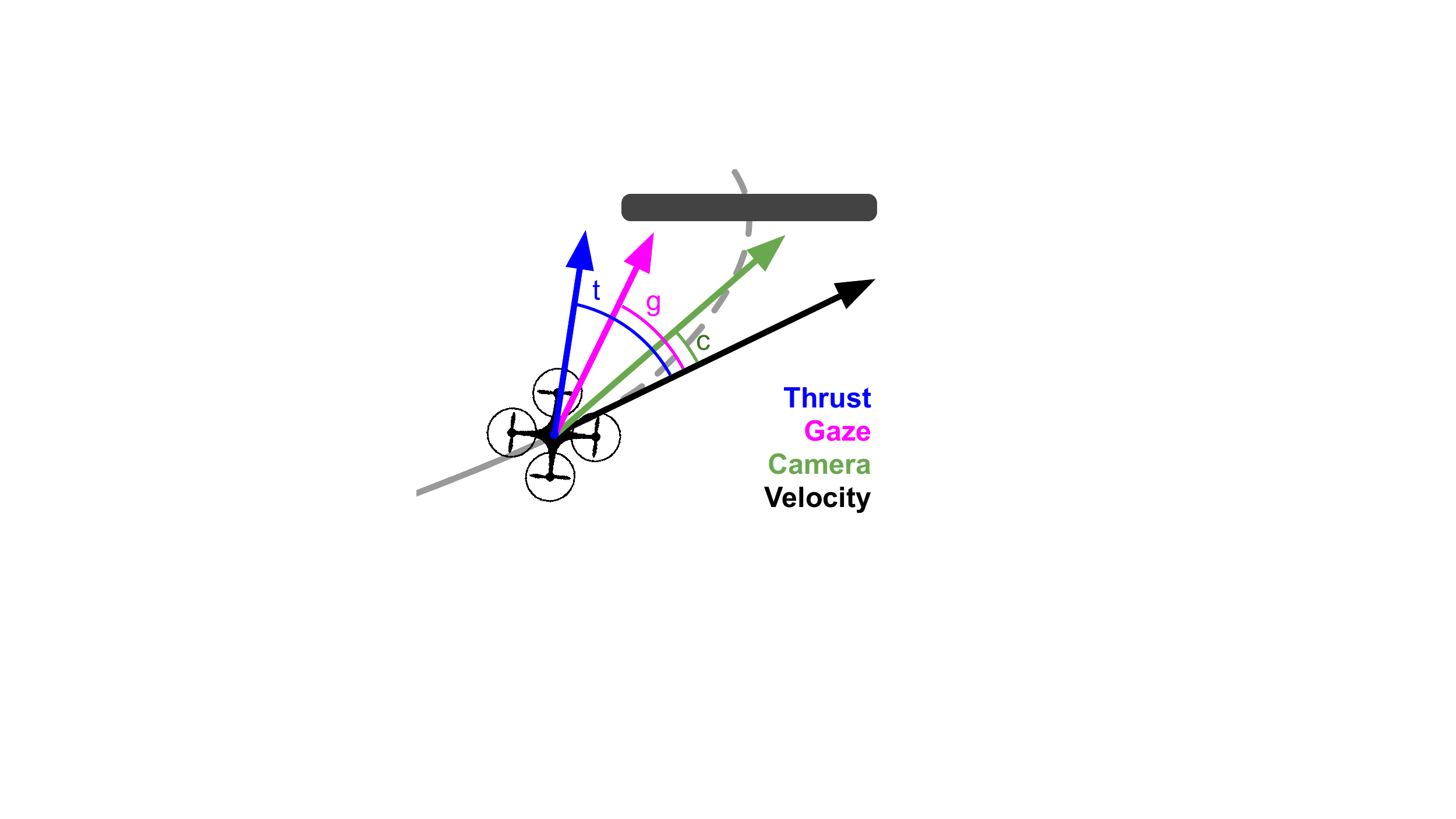}
    \caption{Illustration of vector angle metrics, showing a sche\-matic top-view of a quadrotor approaching a gate (gray horizontal bar) in a left turn maneuver (thin gray line). Arrows represent the thrust vector (in blue), gaze vector (in pink), camera vector (in green), and velocity vector (in black) at the current moment in time. Angle metrics were computed as the signed horizontal angles between each vector and the velocity vector (i.e., t: thrust angle, g: gaze angle, c: camera angle).}
    \label{fig:fig2}
\end{figure}

\subsection{Data Prepossessing}
Data preprocessing consisted of time-synchronization of raw data streams and data quality checks using custom Python scripts. We then processed drone data by converting drone data logs from NED coordinate frame to NWU coordinate frame. We computed additional metrics, such as track progress, i.e. progress in meters along the shortest path trajectory through the gates. This was followed by extracting from raw data the timestamps of gate-passing events and determining whether participants performed valid laps (i.e., all gates passed in the correct order) or invalid laps (i.e., missing gates or invalid order), and detecting collision events from drone acceleration data (i.e. \textgreater $50$ threshold). We then computed lap times from valid laps for further analyses. 

Gaze-to-screen mapping was performed using the surface tracking and undistorting plugins of the PupilCapture software. Twelve fiducial markers with $5$$\times$$5$ cm size were attached to the outside borders of the computer monitor. An offline algorithm detected these markers in the world camera video and computed projection matrices transforming the world to screen coordinates. The world camera image was then undistorted and the resulting projections matrices were computed. These projection matrices were then used to map individual gaze position data from distorted world camera frames to undistorted screen coordinates. Quality control was performed by visual inspection.

Areas of interest (AOIs) were defined as two-dimensional surfaces with the same height and width (no depth) as the gates, placed at the coordinates as gates (see Fig. \ref{fig:fig4}a for an example). We used the Python packages opencv-python and shapely to compute AOI fixations. Ray casting was performed using the camera pose in world coordinates and gaze positions in screen coordinates. This allowed us to compute the gaze vectors in world coordinates (see Fig. \ref{fig:fig2} for an illustration). We then determined whether the gaze vector intersected with an AOI (i.e. AOI fixation event). For each AOI fixation event, we logged the timestamp, gate ID, intersection coordinates, length of the gaze vector from its origin to the AOI intersection point to .csv files for further processing.

Questionnaire data was preprocessed by converting all ratings to $0$-$100$\% scale. 

\subsection{Feature Extraction}
Features to be used for statistical analyses were: ($1$) Drone state: position, norm velocity, norm acceleration and angular rates. ($2$) Control commands: throttle (i.e., collective thrust in Newtons) and body rates (i.e., roll, pitch, and yaw rate in rad/sec). ($3$) AOI first fixations were computed separately for each lap as the time interval between first fixation and passing the AOI (i.e. the gate) with the drone. ($4$) AOI fixation duration was computed separately for each lap as the time interval between first fixation and last fixation of the AOI. ($5$) Distance from fixated object (i.e., AOI or ground floor) was computed for each AOI fixation event as the euclidean distance in meters between drone position and the intersection point between AOI and gaze ray. ($6$) Gaze angle, camera angle, and thrust angle were computed as the signed horizontal angles (in rads) of gaze vector, camera vector, and thrust vector relative to the drone velocity vector (Fig. \ref{fig:fig2} illustrates the relationship between angles and vectors). We note that camera angle was included as a feature, because camera vector is rotated at an offset from the quadrotor z-axis, thus yaw rotations (around quadrotor z-axis) selectively affect camera vector while thrust vector (which is aligned with quadrotor z-axis) is unaffected. Hence, yaw control inputs will differently affect thrust and camera angles.

\subsection{Analysis}
Statistical analyses of flight performance, quadrotor control and collision data were carried out with general linear mixed models with different fixed effects, depending on the goal of the analysis, and a random intercept (Subject). The advantage of these types of modes is that they can be applied on single lap data form a group of different subjects because random between-subjects variance is modeled by the random effect. Therefore these models have more sensitivity of detecting small differences in fixed effects. Cross-correlation analysis was carried out on consecutive time-series for left and right turn sequences of the Flat and Wave tracks across candidate time lags in the interval from $-2$ to $2$ sec (in steps of $1$ ms).

\section{Results}

\begin{figure}[t]
    \centering
    \includegraphics[width=\linewidth,trim={0 0 0 0 },clip]{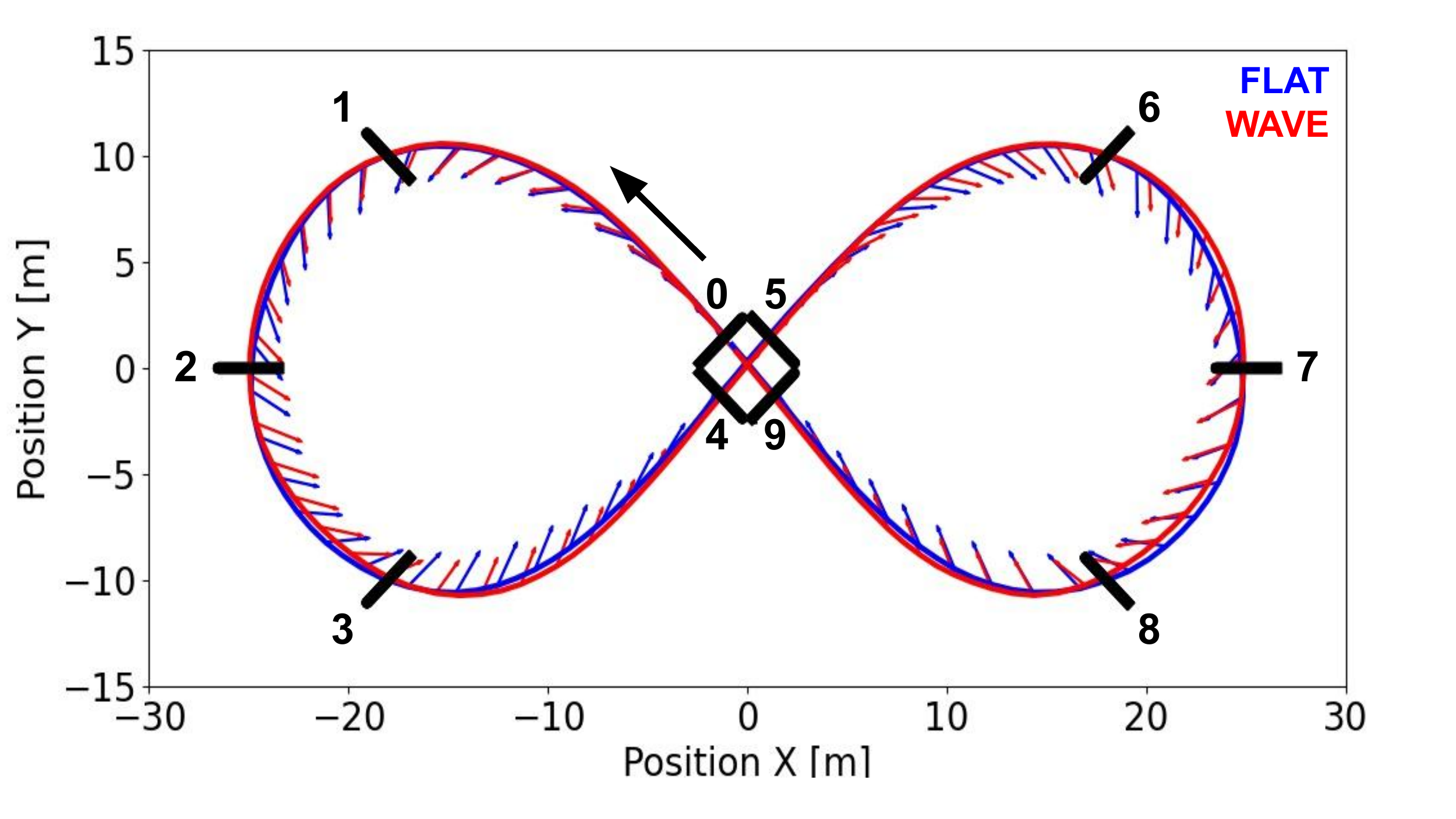}
    \caption{Flight path grand average across participants for the Flat track (in blue) and Wave track (in red). Blue and red arrows indicate the direction of commanded acceleration (thrust) and the black arrow indicates direction of flight. Gate positions are indicated with black lines and gates are sequentially numbered following the direction of flight. }
    \label{fig:fig3}
\end{figure}

\subsection{Flight Performance}
The participants completed a total of $1327$ laps (Flat track: $715$ laps, Wave track: $612$ laps), of which $1277$ laps were valid ($96\%$; Flat track: $680$ laps, Wave track: $597$ laps), i.e. all gates were passed in correct order, and $50$ laps were invalid ($4$\%; Flat track: $35$ laps, Wave track: $15$ laps), i.e. gates were missed or passed in the wrong order. Invalid laps were excluded from further analysis because they were rare and heterogenous (i.e. various crash locations, repeated gate passes, flight in the wrong direction). Among the $1277$ valid laps, $410$ laps ($32$\%; Flat track: $208$ laps, Wave track: $202$ laps) showed collisions events between the drone and a gate ($95$\%) or the ground floor ($5\%$), leading to small perturbations of flight trajectory or crash. For the main analyses of quadrotor control, gaze behavior, and cross-correlations we excluded time periods $-2$ to $2 sec$ relative to collision events. We present a separate analysis of collision data in the Collision Analysis section. 
Flight performance in valid laps was evaluated using the features: number of laps, fastest lap time, maximum velocity, and number of collisions. Because previous work on visual-motor coordination in humans showed time-dependent effects on performance in terms of speed-accuracy trade-off \cite{Heitz2014}, task-specific learning \cite{WORRINGHAM1989}, and vigilance \cite{Russo2006}, we performed a run-by-run statistical analysis to assess the presence of performance changes on performance in our data. The results showed an improvement across runs for number of laps (Run 1 average: $6.53$ laps, Run $5$ average: $7.79$ laps, statistics: T=$4.00$, p\textless$0.0001$), fastest lap time (Run $1$ average: $13.82$ sec, Run $5$ average: $11.93$ sec, statistics: T=$-3.854$, p\textless$0.0001$), and maximum velocity (Run $1$ average: $11.96$ m/s, Run $5$ average: $13.88$ m/s; statistics: T=$7.74$, p\textless$0.0001$). No change across runs was found for number of collisions (Run $1$ average: $21.11$, Run $5$ average: $15.42$; statistics: T=$-1.49$, p=$0.14$). These results indicate an improvement of task-relevant flight performance at a constant level of accuracy across all tested participants. Run-by-run analysis of subjective ratings for overall performance, speed, and accuracy accurately matched the observed performance changes across runs. Task load ratings were modest (i.e., $30-60$\%), and did not indicate that the cognitive demand in drone racing affected flight performances. Fig. \ref{fig:fig3} shows the grand average flight trajectories for the Flat and Wave tracks.

\subsection{Quadrotor Control}
\begin{figure}[t]
    \centering
    \includegraphics[width=\linewidth,trim={60 0 90 0},clip]{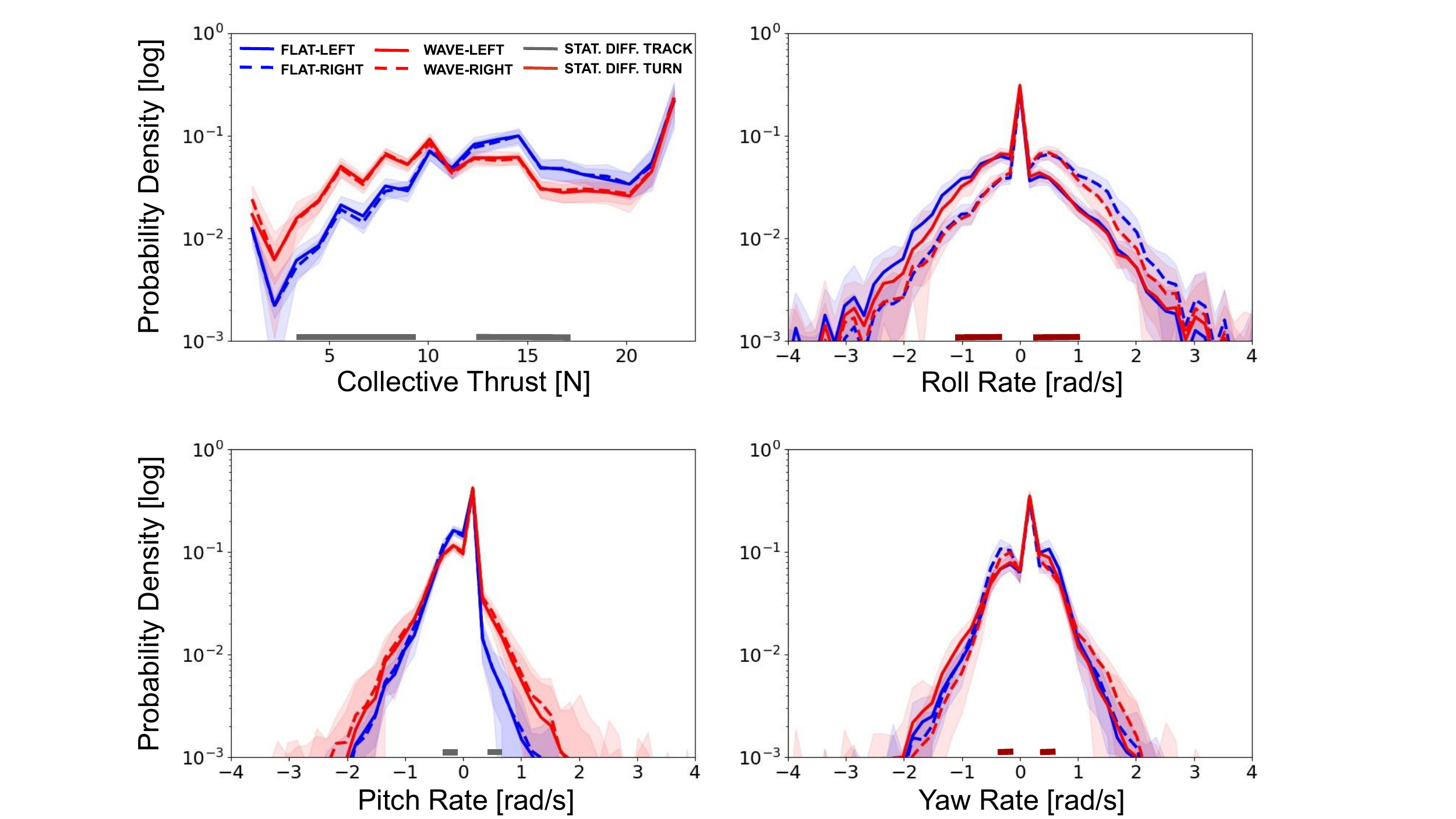}
    \caption{Group-average probability density distributions for commanded thrust and body rates. Data from the Flat track is shown in blue and from the Wave track in red. Left turn sequences (between gates $0-4$) are shown with solid lines and right turn sequences (between gates $5-9$) are shown with dashed lines. Shaded areas show $95$\% confidence intervals. Statistical differences in control command usage between tracks are highlighted in gray and between turns are highlighted in brown.}
    \label{fig:fig4}
\end{figure}

Fig. \ref{fig:fig3} shows the distribution of control commands used by the participants in left turn (gates 0-4) and right turn (gates $5-9$) sequences on the Flat and Wave track. Statistical analysis revealed systematic differences between control command usage for Flat vs. Wave track and left vs. right turn maneuvers. More specifically, participants used more frequently a high throttle input, more upward and less downward pitch on the Flat than the Wave track (Fig. \ref{fig:fig4}a+c). These differences are related to the overall lower velocity and elevation changes required when flying the Wave as compared to Flat track. Moreover, participants used coordinated yaw and roll commands for left vs. right turns (Fig. \ref{fig:fig3}b+d). This pattern of control command usage is to be expected when performing “banked turn” flight maneuvers, in which the  FPV camera is pointed forward while keeping constant altitude throughout a turn.

\subsection{Gaze Behavior}
\begin{figure}[t]
    \centering
    \includegraphics[width=\linewidth,trim={8 0 20 0},clip]{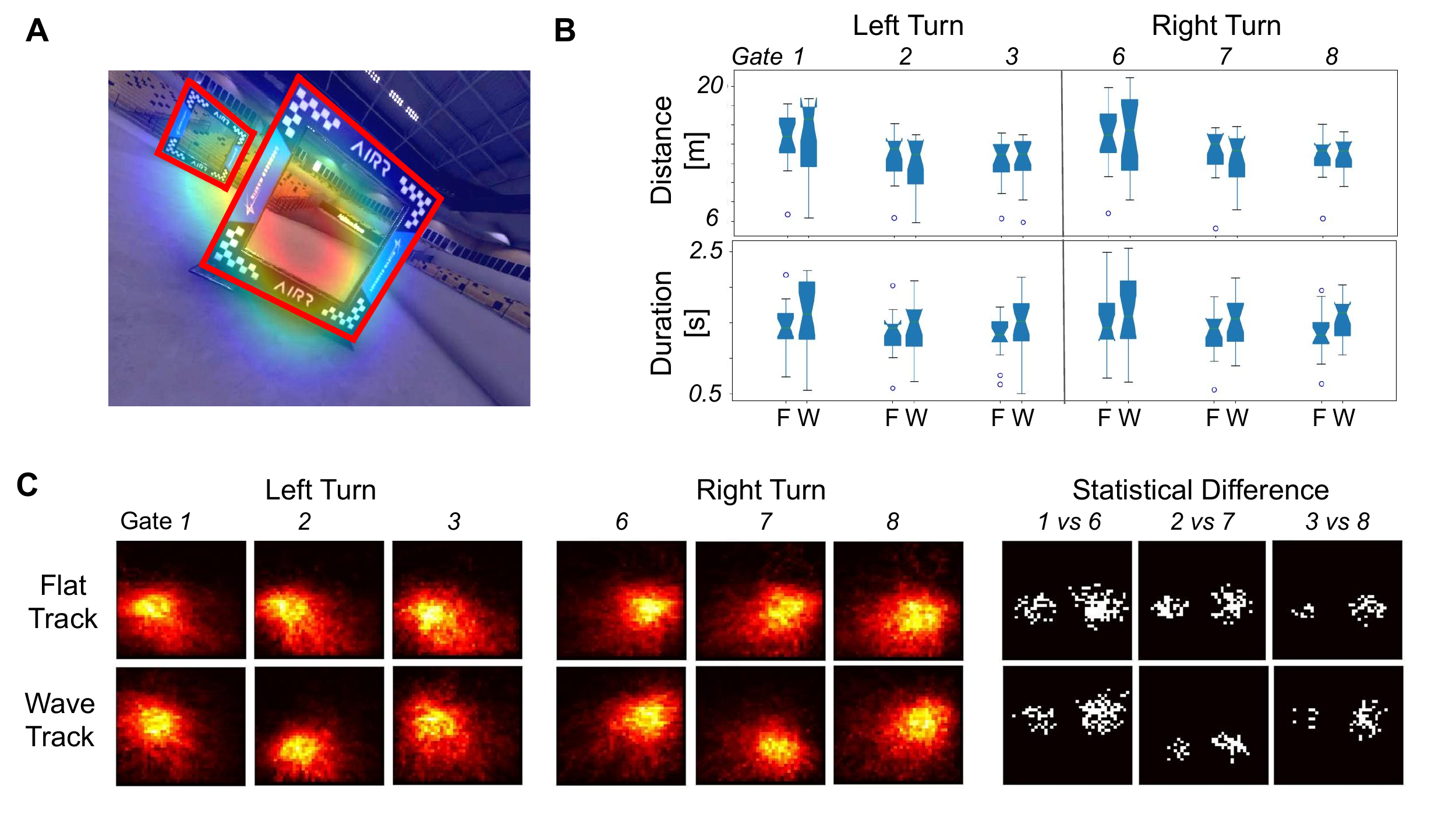}
    \caption{Gaze behavior Area-of-Interest (AOI) analysis results. (A) Illustration of gate AOIs (red outline) and smoothed gaze fixation heatmap for a subject flying the left turn sequence. (B) Boxplots of first-fixation distance from gate and duration until gate pass across subjects for the left and right turn gates of the Flat (F) and Wave (W) track. (C) Grand average AOI fixation probability densities and statistical comparison between left and right turns.}
    \label{fig:fig5}
\end{figure}

Visual inspection of raw gaze data showed that participants fixated their eye gaze most of the time on the upcoming gates. Therefore, we analyzed gaze behavior using AOI analyses (see Fig. \ref{fig:fig5}a for illustration). Fig. \ref{fig:fig5}b shows the distributions of first fixation features across subjects for three gates of the left and right turn sequence of the Flat and Wave track. On average, participants performed the first fixation of gates at $1.5$ sec before reaching the gate, corresponding to an average distance of $16$ meter before reaching the gate. These results indicate that on the figure-eight tracks used in our study participants mainly fixated the next gate and initiated gate fixations as soon as they passed the previous gate. Statistical analyses showed no differences between left vs. right turn and Flat vs. Wave track for the first fixation distance. However, a significant effect of the race track on the first fixation time was found ($p$\textless$.001$). This may be related to the fact that on the Wave track participants flew with an overall lower velocity than on the Flat track due to upward and downward trajectories, which resulted in a slightly longer duration for reaching the gate after first fixation.
Next, we were interested to identify the spatial distribution of fixation locations within the AOIs for different flight maneuvers. We thus extracted probability density distributions of AOI fixations for different gate-passing sequences. Fig. \ref{fig:fig5}c shows the grand average AOI probability density distribution across all subjects for gates of the left and right turn sequence of the Flat and Wave track. Statistical analyses showed a systematic difference between AOI fixation locations for left vs. right turns, indicating that participants fixated the part of the gate that was close to the inner bend of the planned future trajectory. Likewise, sequences requiring an upward-downward trajectory showed AOI fixations on the lower part of the gate, thus also towards the inner bend of the future trajectory.

\subsection{Cross-Correlations}
\begin{figure}[t]
    \centering
    \includegraphics[width=\linewidth,trim={100 0 130 10},clip]{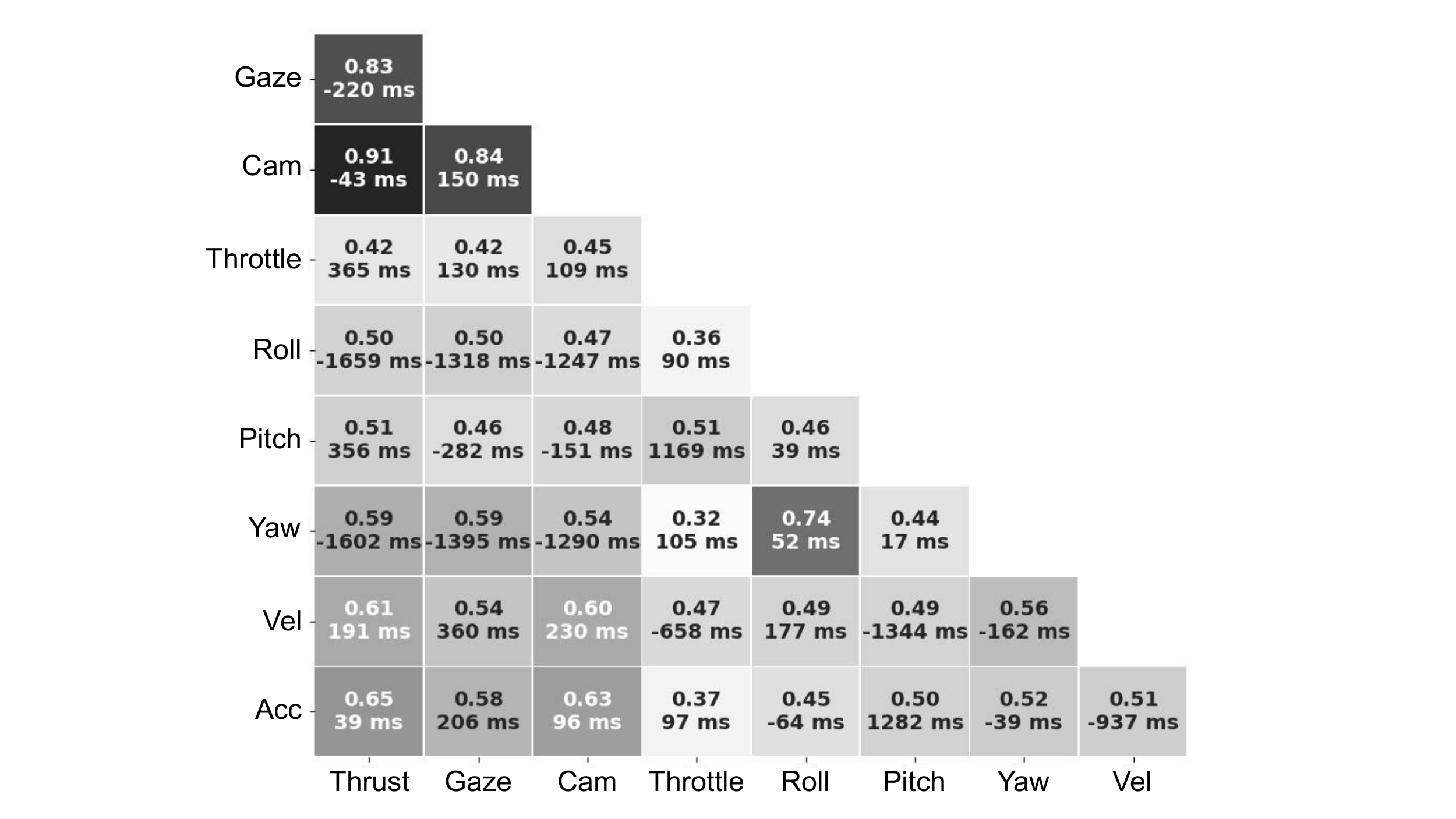}
    \caption{Cross-correlation results for combinations of vector angle (Thrust: thrust angle, Gaze: gaze angle, Cam: camera angle), control command (Throttle, Roll, Pitch, Yaw), and drone state features (Vel: norm. velocity, Acc: norm. acceleration). The first value in each cell refers to the peak correlation coefficient ($r$) and the second value refers to the peak correlation latency in ms. Darker colors refer to higher peak correlation coefficients. Positive latencies indicate that correlated changes in the x-axis variable precede changes in the y-axis variable (negative latencies: y-axis variable precedes x-axis variable).}
    \label{fig:fig6}
\end{figure}

In order to test for the presence of a systematic temporal relationship between multimodal drone state, control command, and eye movement data, we performed cross-correlation analysis. This analysis allows to evaluate the covariation of time-series data at various time lags. We used the peak correlation coefficient r as a metric of the strength of a relationship, and time lag of the peak correlation as a metric of the relative timing between two related signals. Cross-correlation analysis was performed on combinations of the drone state features (i.e., velocity, acceleration), control commands (i.e., throttle, roll, pitch, and yaw rate), and vector angles (i.e., gaze angle, camera angle, and thrust angle. Fig. \ref{fig:fig6} shows an overview of cross-correlation results. We found very strong peak correlations for gaze angle vs. thrust angle (r=$0.83$, $-220$ ms lag), gaze angle vs. camera angle (r=$0.8$, $-150$ ms lag) and camera angle vs. thrust angle (r=$0.91$, $-43$ ms lag), indicating a systematic temporal sequence according to which gaze angle changes precede camera angle changes (i.e. due to yaw input, see Methods for explanation) preceding thrust angle changes. Next, we found a strong peak correlation for roll vs. yaw commands (r=$0.74$, $-52$ ms lag), which corroborates the results from control command analysis regarding the coordinated banked turns performed by our participants. The remaining combinations of drone state, control command and vector angle features showed only moderate cross correlations with peak correlation values ranging from $0.42$ to $0.60$, and temporal lags of up to $1.7$ sec (Fig. \ref{fig:fig6}).

\subsection{Collision Analysis}
Collisions of the drone with gates occurred in approximately $32$\% of the valid laps. Spatial distribution analysis showed that most collisions occurred at the first gate of the left or right turn sequences (i.e., gate $1$: $19$\%, gate $6$: $15$\%), followed by collisions with the last gate of the left or right turn sequences (i.e., gate $3$: $12$\%, gate $8$: $11$\%). The lowest amount of collisions were observed at intermediate and center gates (i.e. $3-8$\%). These differences may be related to the gate rotations relative to the previous and following gates, which for first and third gate required pilots to take a more shallow entry/exit angle as compared to the other gates. Next, we investigated the time periods leading up to collisions. We focused on the $2$ sec time interval leading up to the first collision event of a lap. We randomly selected corresponding time intervals from no-collision laps (i.e., from the same subject and run) and extracted drone state, control command, gaze behavior and vector angle features in the $2$ sec period. General linear mixed models analysis showed differences between collision and no-collision trials for median throttle command (i.e., higher in no-collision than in collision trials, $p$\textless$0.0001$) and for first fixation onset on AOI (i.e. late onset for collision than no-collision trials, $p$\textless$0.01$). No collision-related differences regarding drone state features were found.

\section{Discussion}
This study investigated flight performance, control behavior, and gaze behavior in human-piloted drone racing. Our results showed an improvement of flight performances over time in terms of speed without changes in the overall accuracy, indicating task-specific learning on the race track. These training-related changes were not associated with changes in subjective task demand. Analysis of control behavior showed that pilots coordinated roll and yaw commands for performing banked turns by keeping a forward-facing FPV camera. Although there are numerous differences between drone racing and car driving, we found a strong spatio-temporal relationship between gaze behavior and control behavior similar to previous work in car driving \cite{Marple-Horvat2005PreventionPerformance, Chattington2007Eye-steeringDriving}. 
For instance, \cite{Marple-Horvat2005PreventionPerformance} found a cross-correlation latency difference of $300-400$ ms between eye gaze fixation away from the driving direction of the car and the steering wheel angle. This latency is longer than the $220$ ms observed in our study, which may be related to task-specific requirements, the length and complexity of the chosen flight trajectories, or other factors. However, the observed $220$ ms latency in our study is in line with visual-motor response latencies observed for simple reaction time tasks in humans \cite{Woods2015, Tang2000}. 
One may ask what is the purpose of eye movements in drone racing? Based on previous literature, we propose that eye movements serve for retinal image stabilization for facilitating egomotion estimation \cite{Angelaki2005Self-motion-inducedNavigation, Lappe1999PerceptionFlow}. More precisely, due to the fast quadrotor motion, gate images viewed on the FPV display change position frequently. Smooth pursuit-like gaze fixations allow the observer to center the eye gaze onto the moving stimulus, thereby stabilizing the visual image received by the retina. The remaining visual motion, such as optical flow relative to the gate, and changes of the visual appearance of the gate over time, can now be picked up by brain circuits for egomotion estimation in the visual-parietal cortex \cite{deWaele2001VestibularCortex}. Thus, eye movements may play an important role in supporting state estimation and planning of future control commands. In addition, first target fixations of about $1.5$ sec provide pilots with sufficient time to perform repeated iterations of visual-motor control to eventually successfully pass through a gate.

These results may be highly informative for researchers working on vision-based autonomous navigation. For instance, by studying the eye movement behavior of human pilots one might be able to develop an image stabilization process (e.g. by combining gate detection algorithms with mechanical or digital gimbals) that removes task-irrelevant motion and reduces the negative effects of motion blur on state estimation. In addition, the observed $1.5$ sec and $16$ meters of advance fixations in human pilots may serve to guide the design of receding horizon trajectory planners for drone racing tasks. Finally, the coordinated yaw-roll control behaviors observed in our study may be used for implementing human-like motion planning algorithms that trade visibility of future gates with time-optimal trajectory planning. A potential application of fast vision-based autonomous navigation are search-and-rescue missions. The autonomous drone has to be able to navigate in GPS-denied environments, enter small openings in collapsed buildings, reach survivors as fast as possible, and complete the task within the limited battery life. These capabilities can be tested and benchmarked against the performance of human pilots in drone racing scenarios. A potential limitation of our study is that the observed results may be highly specific to the tested subjects sample of experienced pilots, to the chosen quadrotor model, shape and size of the race track, and racing format. For instance, our work has focused on single player flight, and thus did not require pilots to perform multiplayer related tasks of opponent tracking and collision avoidance. In addition, quadrotor simulators as used here use simplified dynamics and often do not model aerodynamic effects that would be observed in real-world drone racing.

\section{Conclusion}
Our study in human drone racing pilots revealed a strong relationship between eye gaze behavior, quadrotor control, and flight performance. We believe that these results can inspire future developments in vision-based navigation for making autonomous drones faster, more agile, efficient, and safe.

\section{Acknowledgment}
The authors thank Gabriel Kocher for his feedback on the experimental design.


\bibliographystyle{IEEEtran}
\bibliography{main.bbl}

\end{document}